\documentclass[conference]{IEEEtran}
\IEEEoverridecommandlockouts
\usepackage{cite}
\usepackage{amsmath,amssymb,amsfonts}
\usepackage{algorithmic}
\usepackage{graphicx}
\usepackage{textcomp}
\usepackage{xcolor}
\usepackage{url}
\usepackage{hyperref}
\usepackage{todonotes}
\usepackage{tabularx}
\def\BibTeX{{\rm B\kern-.05em{\sc i\kern-.025em b}\kern-.08em
    T\kern-.1667em\lower.7ex\hbox{E}\kern-.125emX}}
\begin{document}

\title{Event-based RGB-D sensing with structured light
}

\author{\IEEEauthorblockN{Seyed Ehsan Marjani Bajestani}
\IEEEauthorblockA{Polytechnique Montreal\\
ehsan.marjani@polymtl.ca}
\and
\IEEEauthorblockN{Giovanni Beltrame}
\IEEEauthorblockA{Polytechnique Montreal\\
giovanni.beltrame@polymtl.ca}
}

\maketitle

\begin{abstract}
  Event-based cameras (ECs) are bio-inspired sensors that asynchronously report
  brightness changes for each pixel. Due to their high dynamic range,
  pixel bandwidth, temporal resolution, low power consumption, and computational
  simplicity, they are beneficial for vision-based projects in challenging
  lighting conditions and they can detect fast movements with their microsecond
  response time. The first generation of ECs are monochrome, but color data is
  very useful and sometimes essential for certain vision-based applications.
  The latest technology enables manufacturers to build color ECs, trading off
  the size of the sensor and substantially reducing the resolution compared to
  monochrome models, despite having the same bandwidth. In addition, ECs only
  detect changes in light and do not show static or slowly moving objects. We
  introduce a method to detect full RGB events using a monochrome EC aided by a
  structured light projector. The projector emits rapidly changing RGB patterns
  of light beams on the scene, the reflection of which is captured by the EC. We
  combine the benefits of ECs and projection-based techniques and allow depth
  and color detection of static or moving objects with a commercial TI
  LightCrafter 4500 projector and a monocular monochrome EC, paving the way for
  frameless RGB-D sensing applications.
\end{abstract}


\section{Introduction}
Event-based cameras (ECs) report brightness changes asynchronously for each
pixel, a behavior inspired by the human eye~\cite{gallego2020event}. When the
brightness changes over a certain threshold for a pixel, the camera generates an
event containing the coordinates of the pixel ($x$,$y$), a timestamp, and the
polarity of the event (i.e. increasing or decreasing). Although ECs do not
capture full images, they can detect movement thousands of times faster than
standard frame-based sensors, and since they do not have an external shutter cycle,
their output is event-driven and frameless, resulting in very low
latency, power consumption, and bandwidth demands.

\begin{figure}[htbp]
	\centerline{\includegraphics[width=0.5\textwidth]{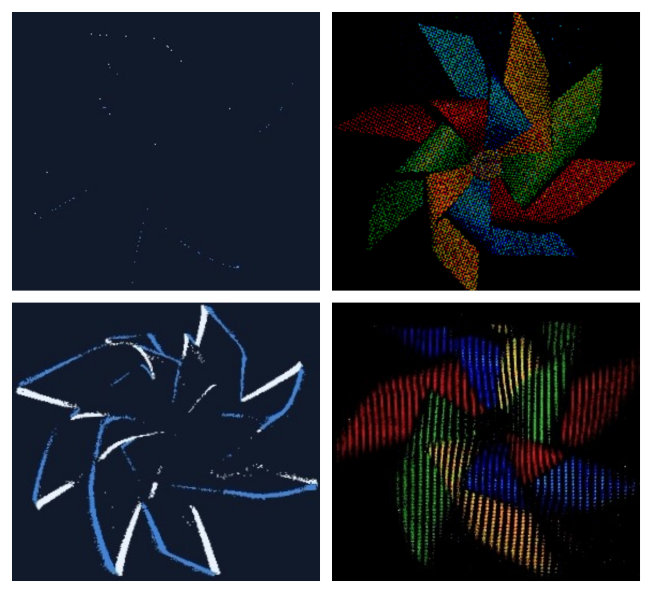}}
	\caption{Color detection of a stable (top row) and spinning (bottom row)
    colorful paper pinwheel. Left column: monochrome events without structured
    light. Right column: colorful image reconstructed aided by structured light
    with two patterns and equivalent speeds of 30 fps (top) and 150 fps (bottom).}
	\label{fig:pinwheel_0}
\end{figure}

Due to their advantages, ECs have been used in various computer vision
applications such as fast movement detection and
tracking~\cite{barrios2018movement, mitrokhin2018event,
  alzugaray2018asynchronous}, optical flow, pose tracking and visual-inertial
odometery~\cite{zihao2017event, mueggler2018continuous}, Simultaneous Localization
And Mapping (SLAM)~\cite{rebecq2016evo, reinbacher2017real},
pattern recognition~\cite{sironi2018hats}, depth estimation and stereo
vision~\cite{zhou2018semi, rebecq2018emvs, steffen2019neuromorphic}, and many more.

In computer vision, color information has an important
role~\cite{tremeau2008color} and could be essential to many tasks such as
segmentation and recognition~\cite{marcireau2018event}. The first generation of
ECs are monochromatic, with color ECs only recently becoming
available~\cite{li2015design, taverni2018front, moeys2017color,
  moeys2017sensitive}. However, due to limitations in terms of sensor size,
color ECs have lower resolution than mono ECs because they need to use color
filters.

It is worth noting that ECs report pixel brightness changes, meaning that an EC
will not report anything when the camera (and/or the object in its field of
view) is static or slowly moving (Fig.~\ref{fig:pinwheel_0}, top left), which
can be critical in some cases (e.g. for a slow-moving robot). 
To overcome this issue, one could use an external active device such as a laser,
a flashing LED, or a light projector to generate events in static and almost
static situations. This external active lighting system could also be used to
detect depth by projecting detectable patterns called Structured Light (SL)~\cite{brandli2014adaptive,
  matsuda2015mc3d,
  muglikar2021event, muglikar2021esl}.

We present a method to add color and depth to a monocular, monochrome
event-based camera while maintaining fast response time and resolution. We use a
Digital Light Processing (DLP) projector that emits patterns of lights that we
call Active Structured Light
(ASL) 
on a scene, the reflection of which is captured by the EC which in turn
generates events tagged with the color and depth of the scene. It is worth noting
that our ASL method could also be used with color ECs, allowing the detection
of static scenes. By dynamically adjusting the projection, we have color data
when needed, managing the overall bandwidth of the system. For example, we can
use the full resolution of the camera to detect static color scenes, or use more
sparse patterns for fast moving objects. Projecting patterns also allows
triangulation-based measurements to create a colorful 3D point cloud of the
scene.

Overall, we present 5 contributions:
\begin{enumerate}
	\item the generation of colorful events from a monochrome EC for various speeds of movement;
	\item a higher-resolution version of a color EC using a monochrome EC;
	\item we allow the detection of static objects and scenes;
	\item we optimize the bandwidth usage of the EC by detecting the color when and where it is needed;
	\item we use patterns that allow event-based depth measurement, ultimately
    generating colorful point clouds.
\end{enumerate}

\begin{figure}[htbp]
	\centerline{\includegraphics[width=0.5\textwidth]{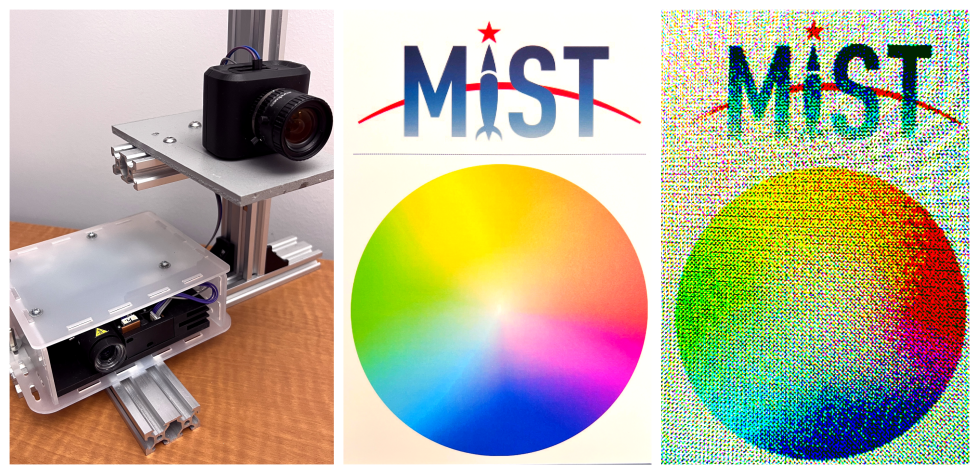}}
	\caption{Left: The experimental setup with a DLP LightCrafter 4500 Evaluation
    Module and a Prophesee evaluation kit (Gen3-VGA). Middle: Printed color
    wheel with the logo of the \href{https://mistlab.ca/}{MIST Lab.}, captured
    by a frame-based high-resolution camera. Right: Colorful image reconstructed
    by proposed method captured by monochrome EC aided by SL.}
	\label{fig:setup}
\end{figure}

In this work we focused on visual light wavelength (emitted by the LED
projector) and materials that are not in the category of fluorescence and they
do not change the wavelength of the light. We validated our approach in
different dynamic conditions: Fig.~\ref{fig:setup} shows the experimental setup
with a DLP
projector\footnote[1]{\href{https://www.ti.com/tool/DLPLCR4500EVM}{LightCrafter
    4500 Evaluation Module}} and a Prophesee evaluation
kit\footnote[2]{\href{https://docs.prophesee.ai/2.2.2/hw/evk/gen3.html}{Gen3-VGA}}.
With this setup, we achieved full color detection at an equivalent rate of 1400
frames per second (fps) (note that the camera is frameless, we use fps just for
the purpose of comparison). Fig.~\ref{fig:setup} also shows the color detection
of a static printed color wheel.

The rest of the paper is as follows: Section~\ref{rel_works} presents related
work; Section~\ref{mono2rgb} describes our method for color detection;
Section~\ref{acls} details the results of our method in several conditions; and
finally, Section~\ref{conclusion} draws some concluding remarks and outlines
possible future work.

\section{Related Work}\label{rel_works}
Digital color cameras use various Color Filter Array (CFA) or Color Filter
Mosaic (CFM) on their sensors to detect different colors for each pixel, and
among them, the Bayer array filter~\cite{bayer1976color} is the most common
CFA~\cite{ramanath2005color}. Using a CFA, the resolution of a monochrome
photosensor decreases dramatically, since the size of the CFA could be between 4
to 36 pixels 
or even bigger based on its design~\cite{khashabi2014joint}. This means that,
even with the smallest CFA, to have a color pixel we need several monochrome
pixels, which decreases the available resolution (e.g. 4 times with a CFA size
of $2\times2$ pixels).

Colorization is the process to generate a color image based on a monochrome
sensor or grayscale image without loss of resolution. Colorization requires
either external data about the image colors, user interaction, or a trained
neural network embedding the knowledge of the colors on the scene, and can be a
time-consuming and expensive task~\cite{levin2004colorization}. Levin et
al.~\cite{levin2004colorization} introduced a method that needs a few initial
inputs from a user to generate a full color image and keeps tracking the color
on upcoming frames in a video. Zhang et al.~\cite{zhang2016colorful} introduce a
fully automatic colorization approach based on a convolutional neural network
(CNN) that can change a grayscale image into a near-real colorful image. Their
method successfully deceived 32\% of human participants in distinguishing the
generated image compared with the ground truth image. In contrast to these
colorization approaches, our method does not need initial input
data to get color out of a monochrome camera and it can provide realistic
color information faster than CNN models.

Another approach to generate color data without quality loss on a monochrome
image is to use separate cameras: the monochrome sensor takes a more
detailed and higher contrast image, while a lower resolution RGB camera adds
color information. This combination of an RGB camera with a monochrome sensor is very
common, but the image fusion, colorization, or the color transfer process
are still a challenge~\cite{jang2020deep}.

Event-based cameras have introduced a new field of imaging systems. Due to their
advantages compared to standard cameras, many scientists investigated ways to
generate and reconstruct images from events to use in frame-based computer
vision algorithms. A monochrome EC has been used in many image reconstruction
works~\cite{munda2018real, scheerlinck2018continuous, rebecq2019events,
  scheerlinck2019asynchronous, pan2019bringing, haoyu2020learning}. Also, a
dual-cameras consisting of a standard frame-based camera and an EC can produce
a deblurred high frame rate (HFR) and high dynamic range (HDR)
video~\cite{messikommer2022multi}.

By combining three ECs using dichroic filters Marcireau et
al.~\cite{marcireau2018event} introduced a prototype to capture a stream of
events in RGB separate channels for color segmentation. This allowed them to
keep the resolution of the monochrome EC, however, the required bandwidth
increased by three times.

With the introduction of color event-based cameras~\cite{moeys2017color}, some research
focused on the reconstruction of images and videos based on color
events~\cite{rebecq2019high, pan2020high, mostafavi2021learning}. Scheerlinck
\emph{et~al}.~\cite{scheerlinck2019ced} presented a dataset for color ECs. They
also compared the output quality of some image reconstruction methods such
as~\cite{scheerlinck2018continuous, munda2018real, rebecq2019events} in color.

As digital color cameras, current color ECs also use CFA to generate color
events, which reduces their output resolution. Compared to the method presented
by Marcireau et al.~\cite{marcireau2018event}, despite having a lower
resolution, the CFA-bassed color ECs need lower bandwidth. Our method
reconstructs color data when it is needed, which controls the bandwidth of the
system.

\section{Monochrome to color}\label{mono2rgb}
Compared to frame-based cameras, ECs are faster sensors, however, since they
report nothing in a static situation or with slowly moving objects, they require
an additional sensor to provide visual perception in these situations. We use an
external event generator, namely a DLP projector. By emitting a pattern of
light on objects in the scene, not only we can detect their color, but we are
also able to detect depth, which makes event-based RGB-D sensing possible.
Moreover, since ECs have a high dynamic range sensor, there is no need to have a
high-power light projector to see dark environments.

There are many standard color formats for digital color descriptions (additive
or subtractive), such as CMY (cyan, magenta, yellow), or with black CMYK, RYB
(red, yellow, blue), RGB (red, green, blue) or with white RGBW and etc.
Selecting the color space could depend on the application and the color range of
the desired objects in the environment. Without loss of generality, we select
the RGB color space which is more common in vision applications.

\begin{figure}[htbp]
	\centerline{\includegraphics[width=0.5\textwidth]{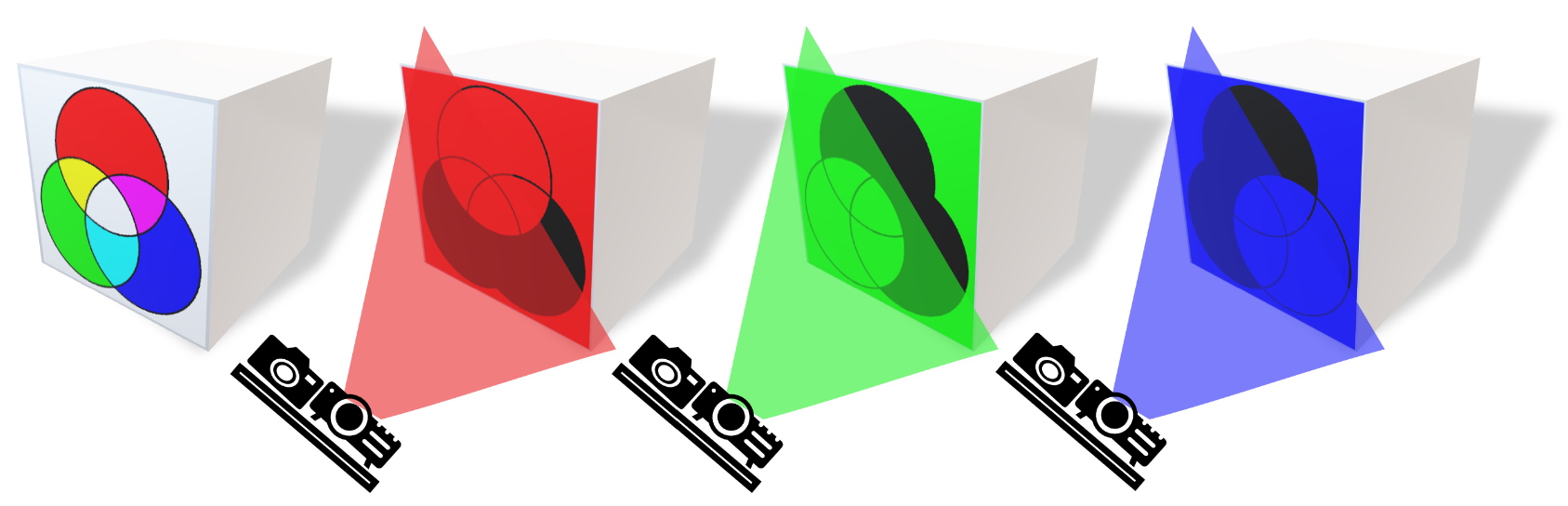}}
	\caption{The procedure of detecting color with a monochrome event-based camera
    aided by structured light. From left to right: The colorful object, SL
    pattern projected in red, green, and blue by the DLP projector and
    collecting the events by EC.}
	\label{fig:rgb_collector}
\end{figure}

We use the EC to measure the amount of reflection of the emitted light on
an object. To measure the color, we project three different wavelengths
(structured light in red, green, and blue) on the environment and measure the
amount of reflected light captured by the EC. During each pattern exposure time,
the received events are gathered in an appropriate color channel on the initial
frame. Fig.~\ref{fig:rgb_collector} shows the procedure for color detection.

\begin{figure}[htbp]
	\centerline{\includegraphics[width=0.38\textwidth]{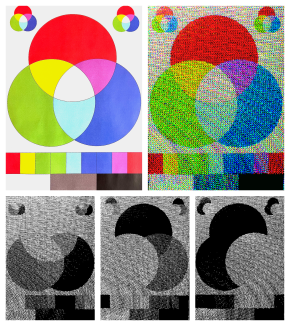}}
	\caption{Color detection of a printed color wheel. Top left captured by
    frame-based high-resolution camera, top right is colorful image
    reconstructed by proposed method, captured by a VGA monochrome EC aided by
    SL. Bottom from left to right are collected event-frames for each color
    light (red, green and blue) by monochrome EC.}
	\label{fig:rgb_colorwheel}
\end{figure}

To synchronize the DLP projector with the EC, we connect the trigger pins of the
camera to the projector. By changing the pattern color, the DLP sends a pulse to
the camera which identifies the incoming events as belonging to the appropriate
color channel. Fig.~\ref{fig:rgb_colorwheel} depicts the output of the color
detection of a printed RGB color wheel separated in each color channel. The
bottom frame of the Fig.~\ref{fig:rgb_colorwheel} shows that the printed color
wheel does not have pure green $(0,255,0)$ and blue $(0,0,255)$ colors in 24 bit
RGB format. For example, in the red light channel (bottom left), the green
circle also reflected some light (although less than the red circle) and as a
result, it appears gray.

\subsection{Color detection speed limits}
\begin{figure}[htbp]
	\centerline{\includegraphics[width=0.5\textwidth]{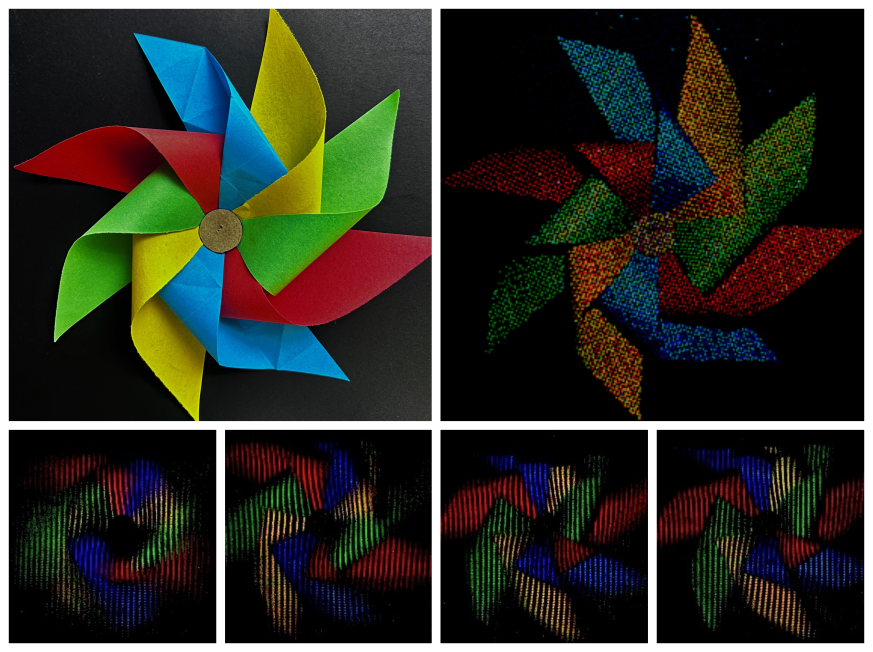}}
	\caption{Color detection of a spinning colorful paper pinwheel reconstructed
    at different frame rates. Top row (static) left: captured by
    frame-based high resolution camera, right: colorful image reconstructed by
    proposed method captured by monochrome EC. Bottom row (spinning pinwheel)
    reconstructed at, from left to right, 30, 100, 120 and 150 fps.}
	\label{fig:pinwheel}
\end{figure}

One of the main advantages of the ECs is their response time which is in the
range of microseconds. However, with the introduced method, we need to gather
events of each color separately, limiting the speed of color detection to
the maximum speed of pattern switching of the DLP projector. With the
LightCrafter 4500 Evaluation Module, we are able to detect color with an
equivalent frame rate up to 1400~fps due to its high frequency
(4225~Hz\footnote[1]{\href{https://www.ti.com/lit/pdf/dlpu011}{Switching rate
    for preloaded 1 Bit depth pattern of the LightCrafter 4500 Evaluation
    Module}}). However, assuming that the color of the object is not changing,
we could still use the other methods to track the object only based on the high
speed stream of events~\cite{mitrokhin2018event, alzugaray2018asynchronous} and
use the color detection method for a short period of time.
Fig.~\ref{fig:pinwheel} shows the output of the color detection of a spinning
colorful paper pinwheel reconstructed at different frame rates.

\subsection{Advantages over monochrome frame-based cameras}
Monochrome or grayscale cameras have been used in vision-based applications that
do not need color information. As mentioned in Section~\ref{rel_works},
the combination of a monochrome camera and a color camera could be challenging
for image fusion, colorization, or the color transfer process~\cite{jang2020deep}. A
dual-camera consisting of a frame-based camera and an EC can produce a
deblurred high frame rate (HFR) and high dynamic range (HDR)
video~\cite{messikommer2022multi}. However, by adding a second camera the
required bandwidth increases at the same time. Our method
allows us to benefit from ECs' features and detect/update the color
information for a given period of time. Moreover, the
camera-projector combination enables depth measurement and makes the
feature detection and matching easier (compared to stereo
cameras)~\cite{brandli2014adaptive,
	matsuda2015mc3d,
	muglikar2021event, muglikar2021esl}.

\subsection{Advantages over color event-based cameras}
As mentioned in Section~\ref{rel_works}, digital cameras often use CFA to detect
color. For instance, the Color-DAVIS346~\cite{taverni2018front} is one of the
most recent color EC that uses an RGBG Bayer pattern with an output resolution
of $346 \times 260$ pixels. This kind of camera is reporting the stream of
events in 3 or 4 different channels, which increase the need for bandwidth.
Higher bandwidth requirements can cause bus saturation (as described in
Section~\ref{acls}). In addition, despite increasing the bandwidth needs and
decreasing the resolution, color ECs cannot detect the environment when the
camera or object is static or moving very slowly. Our method is useful to
efficiently use the bandwidth by detecting the color only when and where it is
needed. Moreover, our method also gathers information from the environment from
an initially static robot or camera, meaning there is no need to have mechanical
parts to move the camera and receive events, which makes the system more
reliable. Further, since a high-resolution EC could be subject to more noise in a dark
environment compared with a low-resolution EC~\cite{gehrig2022high}, our method
could still get the benefits of the high-resolution monochrome EC in a dark
environment.

\subsection{White balance and color correction}\label{white_balance}
White balance and color correction can make the captured image close to its
natural color. White balance can be adjusted before or after capturing the
image. Generating white light with the DLP projector can change the image
white balance, because the color temperature of a light source or the
warmth/coolness of the white light can change the white balance directly. The
DLP projector has three different LED colors: red, green, and blue. Generating
LED-based white light could be challenging with wideband wavelength RGB
LEDs~\cite{muthu2002red,muthu2003red,david2018led}. Since the DLP projector has
narrowband LEDs, changing the white balance can be done relatively easily by changing
the current of each LED separately. 

\textbf{Lighting model:} If we consider the DLP projector as a point-sized light
source, we can model the lighting with the Lambertian shading model which is
one of the simplest bidirectional reflectance distribution functions (BRDF) and
an appropriate approximation to many real-world material
surfaces~\cite{pharr2016physically}. In the Lambertian shading model, R, G, B values
of the resulting pixel are independent of the angle that the viewing ray hits the
surface: 
\begin{equation}
W = S_{ref}S_{pow} max (0, n\cdot b) \label{lambertian_w} ,
\end{equation}
where $W$ is the combination of $(R, G, B)$ values for a desired pixel, and
$S_{ref}$ is the spectral reflectance of the material, $S_{pow}$ represents the
spectral power distribution of the projector (as the light source), $n$ is the
outward surface normal (of the object) and $b$ is the light beam vector which is
from the surface intersection point to the projector. The dot product of these
two unit vectors gives the amount of attenuation based on the angle between the
surface to the projector. The $max$ function is used to prevent a condition
where $n\cdot b<0$, because the projector would be behind the object in this
case. This model could be divided for each color, for example the model for red
light is:
\begin{equation}
R = S_{ref_{R}}S_{pow_{R}} max (0, n\cdot b) \label{lambertian_r}
\end{equation}
To generate white light in an ideal situation, we consider that each color
has the same power distribution and $S_{pow_{R}} = S_{pow_{G}} = S_{pow_{B}}$.
And for a white or gray surface we would have $S_{ref_{R}} = S_{ref_{G}} =
S_{ref_{B}}$. As a result, by controlling the current of each LED ($S_{pow}$) we
can have balanced white light.

\begin{figure}[htbp]
	\begin{center}
    \setlength{\tabcolsep}{1pt}
	\begin{tabular}{ccc}
			\includegraphics[width=1.1in]{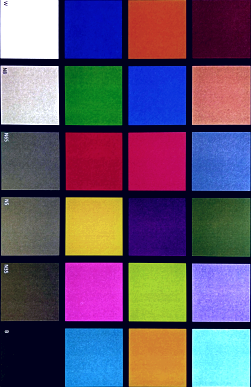}&
			\includegraphics[width=1.1in]{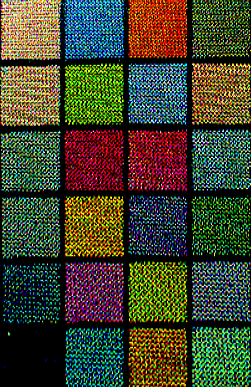}&
			\includegraphics[width=1.1in]{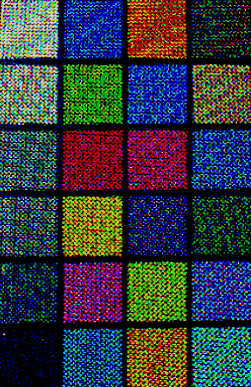}\\
			GT&$HC=0.22$&$HC=0.76$\\
	\end{tabular}
	\caption{Color detection of printed Macbeth color chart. Image captured by a
    frame-based high-resolution camera (left), the colorful image reconstructed
    by the proposed method captured by monochrome EC, without (middle) and with
    (right) white balance.}
	\label{fig:macbeth_chart}
	\end{center}
\end{figure}

We can use a gray card, a color wheel, or a Macbeth color chart/color checker to
calibrate our system. We used a printed Macbeth color chart to do the
calibration, and Fig.~\ref{fig:macbeth_chart} shows the output of the color
detection with the proposed method with and without white balance calibration.

\textbf{Absolute error:} To check the quality of the
reconstructed image, we need to have a base image and specify an error
calculation method. We consider the captured image by a frame-based
high-resolution camera as the base image (Ground Truth or GT) in
Fig.~\ref{fig:macbeth_chart}. To calculate the absolute error, we compared the
histogram of two images in Hue Saturation Value (HSV) format based on the
correlation metric\footnote[1]{\href{https://docs.opencv.org/3.4/d6/dc7/group_imgproc_hist.html}{The
    OpenCV histogram comparison correlation method.}}:

\begin{equation}
c=d(H_{o},H_{b})=\frac{\sum_{I}(H_{o}(I)-\bar{H_{o}})(H_{b}(I)-\bar{H_{b}})}{\sqrt{\sum_{I}(H_{o}(I)-\bar{H_{o}})^{2}\sum_{I}(H_{b}(I)-\bar{H_{b}})^{2}}} \label{correlation} ,
\end{equation}

where 

\begin{equation}
\bar{H_{k}}=\frac{1}{N}\sum_{J}H_{k}(J) \label{histo} ,
\end{equation}

and $N$ is the total number of histogram bins, which in our case is
256 (8 bit in each color channel). %
The $H_{o}$ and $H_{b}$ are respectively histogram of
the output image and the baseline image with a Histogram Correlation ($HC$)
between 0 and 1. The $HC$ between the base image (left) and each reconstructed
image is respectively $0.22$ and $0.76$ for the reconstructed image without
white balance (middle) and with white balance (right) in
Fig.~\ref{fig:macbeth_chart}. Moreover, to check the difference between each
pixel in the reconstructed image and the GT, and calculate the absolute error,
we calculated the root mean square error (RMSE) separately for each channel. As
an example, the RMSE for the red channel is:
\begin{equation}
RMSE_{r}= \sqrt{\frac{\sum_{N}(p_{o_{r}}-p_{b_{r}})^{2}}{N}} \label{rmse} ,
\end{equation}
where $p_{o_{r}}$ and $p_{b_{r}}$ are respectively the pixel value in the red
channel of the output frame and the baseline frame. $N$ is the total number of
pixels, i.e. $640\times480 = 307200$. Table~\ref{table:base} shows the quality
of the color detection for each image in Fig.~\ref{fig:macbeth_chart} compared
to the GT image. Table~\ref{table:base} also shows that, after manual white
balance tuning, all three channels had $12\%$ better RMSE on average. To have a
more realistic color detection, an online white balance calibration could be
helpful in minimizing the average RMSE if needed. To check the quality of each
image we also calculated their Peak Signal-to-Noise Ratio (PSNR), also shown in
Table~\ref{table:base}.

\begin{table}[htbp]
	\caption{Color detection quality table compared to the ground truth~(GT) frame}
	\begin{center}
		\begin{tabular}{|l|c|c|c|}
			\hline
							& \textbf{GT}	& \textbf{No WB}	& \textbf{WB}\\
            \hline
			$RMSE_{red}$	& 0				& 83.89				& \textbf{83.65} \\
			$RMSE_{green}$	& 0				& 87.79				& \textbf{71.16}\\
			$RMSE_{blue}$	& 0				& 92.76				& \textbf{76.97}\\
			$RMSE$			& 0				& 88.15				& \textbf{77.26}\\
			$PSNR$			& 0 dB			& 9.88 dB			& \textbf{8.82 dB}\\
			Histogram Correlation (HC) & 1		& 0.22				& \textbf{0.76}\\
			\hline
		\end{tabular}
		\label{table:base}
	\end{center}
\end{table}

Another way to do the color correction is to capture the image with a white
channel by RGBW color spaces and perform the correction on four channels similar
to RGBW CFA-equipped sensors~\cite{choi2020color}. This comes at the cost of
adding a 4th color light to the SL, adding at least $33\%$ to the length of the
capture time.

\section{ASL: Adaptive Structured Light}\label{acls}
\begin{figure}[htbp]
	\centerline{\includegraphics[width=0.5\textwidth]{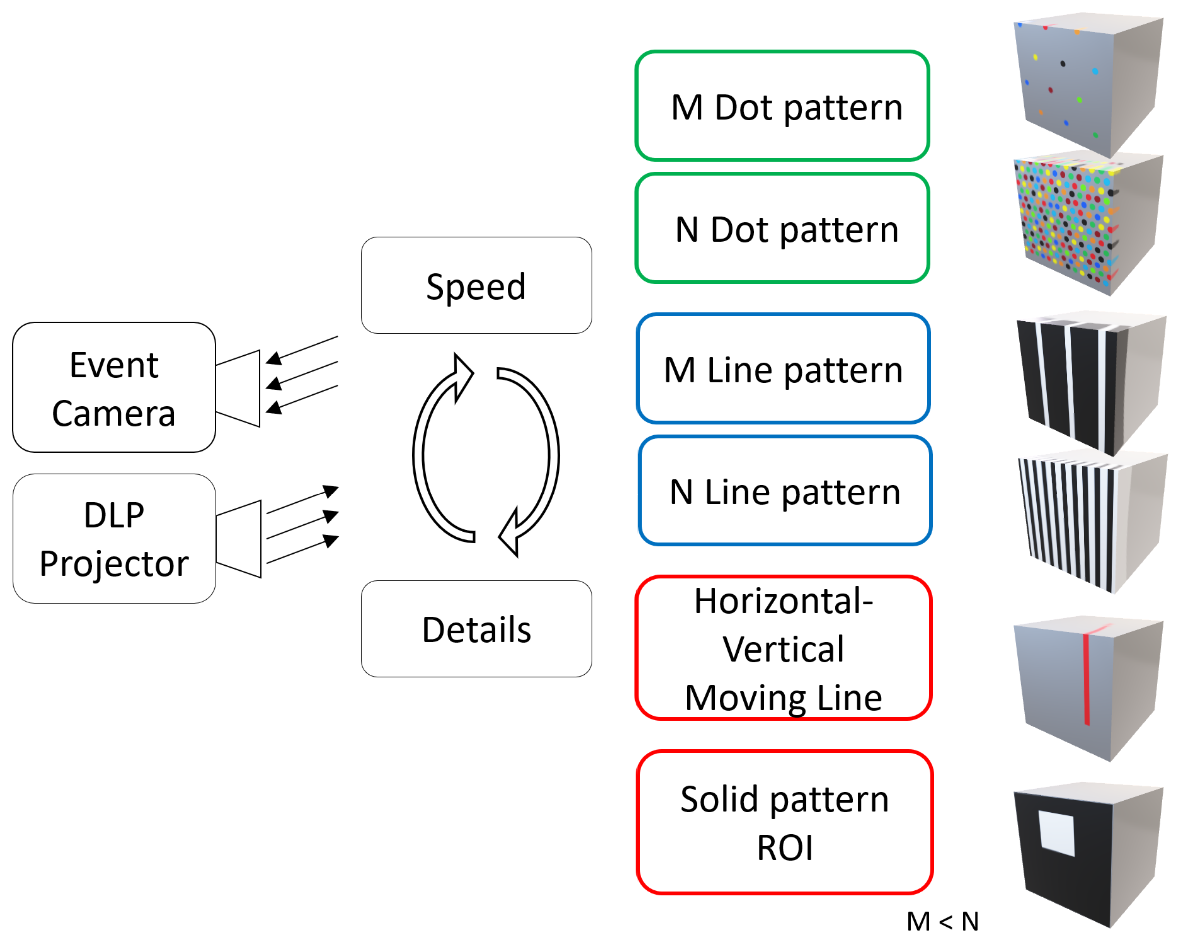}}
	\caption{Different types of SL patterns have been used to control the event rate by Adaptive Structured Light.}
	\label{fig:changing_SL}
\end{figure}

High-resolution ECs have a higher event rate and need more bandwidth compared to
low-resolution ECs, but each EC has a limited data rate (finite bandwidth) on
the output interface or bus. If the data rate or the number of events exceeds
the limit, bus saturation could happen~\cite{gallego2020event,gehrig2022high}.
Filtering~\cite{finateu20205} or online event-rate
control~\cite{delbruck2021feedback} can mitigate this issue. When using an
external event generator such as the DLP projector which emits SL on the scene,
controlling the event rate is even more important. One method to control the
event rate when using a projector is to define a region of interest (ROI) and
project the pattern only where it is needed. Muglikar et
al.~\cite{muglikar2021event} used one EC camera to detect the ROI (generally the
area of the image frame that has more events due to the movement) and then
projected the SL on that area followed by detecting the depth with a second EC.
Instead of adding a second EC to the system, we introduced ASL to control the
event-rate. Fig.~\ref{fig:changing_SL} shows different patterns of the SL which
change based on the number of received events. As expected, there is a trade-off
between having high-resolution (dense) and high-speed (sparse) color
detection. The generated SL patterns are:
\begin{enumerate}
	\item Dot pattern
	\item Multiple-lines pattern
	\item Moving-line pattern
	\item Solid pattern
\end{enumerate}
In static conditions, ASL could also be used with a color EC and white light.
However, it should be noted that color ECs need more bandwidth compared to
monochrome ECs with the same resolution.

\textbf{Bandwidth control:} By frequently projecting SL into the scene, we
receive events caused by the SL alongside the events caused by the movement of
objects or the camera. We want to control the biases of the camera to prevent
bus saturation, but we do not want to lose data by excessively decreasing the
camera sensitivity. In general, the total number of
events should be less than the maximum bandwidth of the EC:
\begin{equation}
Max. Bandwidth < events_{SL} + events_{M}\label{bandwidth},
\end{equation}
where $events_{M}$ is the number of events caused by the movement of the EC or
any object in the scene (i.e., any other events that have not emerged due to the SL).
$events_{SL}$ is the number of events caused by the SL, and we can
control it by changing the pattern and the power of the LED projector.

$events_{SL}$ is not only linked to the color of the object (and its
reflectivity/fluorescence percentage, which we do not investigate in this
paper), but also it is related to the distance of the camera-projector from the
object. Increasing the distance, the spectral power distribution decreases
because of the reduction in power density. Unfortunately, there is no
information available concerning the variation of power density changes with
distance for each LED of the DLP projector. Modelling the DLP projector power
density could be useful, but it is out of the scope of this paper.
In this work, we make the simplifying assumption that the power density of all
LEDs is the same. As a result, to control $events_{SL}$ at a same distance,
we need to control $S_{pow}$ from \eqref{lambertian_w}.

Considering a one-bit pattern, we can control $S_{pow}$ by changing the
pattern (changing the number of white pixels in a black and white frame),
instead of changing the current of the LEDs. We call the number of white pixels
per frame as the coverage percentage (CP), with each pattern type having a different CP.
To additionally simplify the problem, we assume that the DLP and the EC are
close and we can consider the CP on the DLP frame
plane despite the fact that, depending on the relative pose of the camera to the
projector, the CP could be different on the camera frame plane.

\begin{figure}[htbp]
	\centerline{\includegraphics[width=0.4\textwidth]{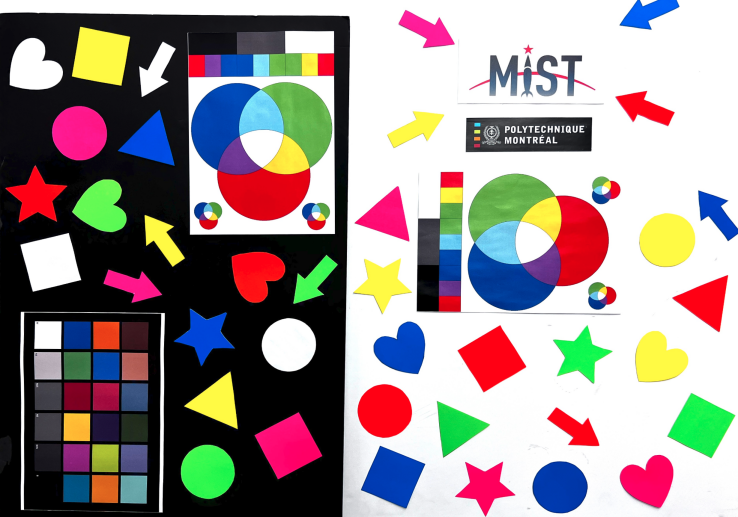}}
	\caption{A colorful board used to test ASL. The board was
    placed at a distance of 160 cm from the camera projector.}
	\label{fig:ACSL_setup}
\end{figure}

Fig.~\ref{fig:ACSL_setup} shows the colorful board that we use in our
experiments, placed $160cm$ from the camera-projector.

\subsection{Dot pattern}
\begin{figure*}[htbp]
	\begin{center}
		\renewcommand{\tabcolsep}{1pt}
	\begin{tabular}{ccc}

		\includegraphics[width=0.32\textwidth]{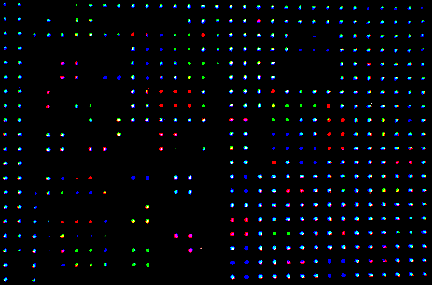}&
		\includegraphics[width=0.32\textwidth]{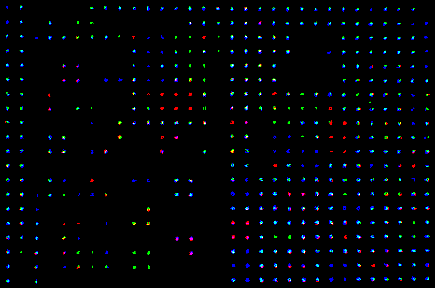}&
		\includegraphics[width=0.32\textwidth]{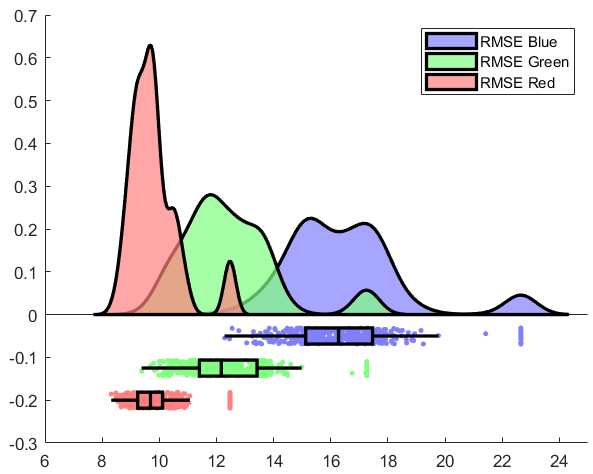}\\
		CP=1.54\%	& RMSE=10.78			& Avg. RMSE=12.94\\
		\hline
		\includegraphics[width=0.32\textwidth]{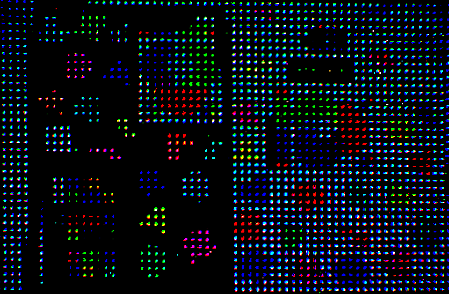}&
		\includegraphics[width=0.32\textwidth]{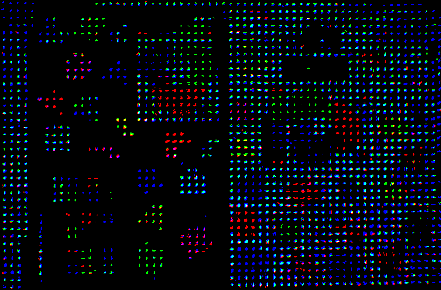}&
		\includegraphics[width=0.32\textwidth]{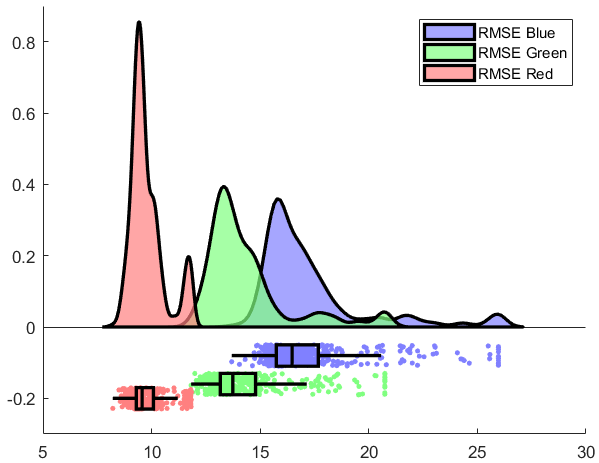}\\
		CP=2.22\%		&RMSE=11.34			& Avg. RMSE=13.83\\
		\hline
		\includegraphics[width=0.32\textwidth]{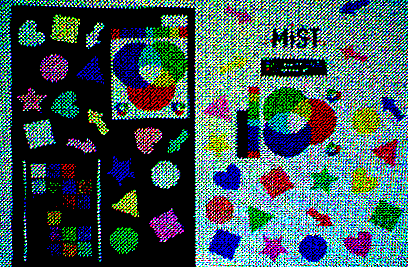}&
		\includegraphics[width=0.32\textwidth]{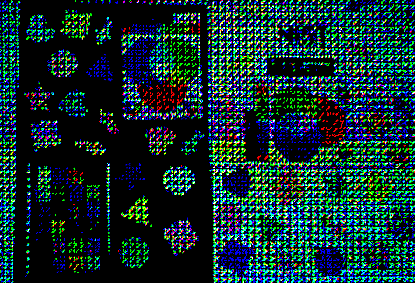}&
		\includegraphics[width=0.32\textwidth]{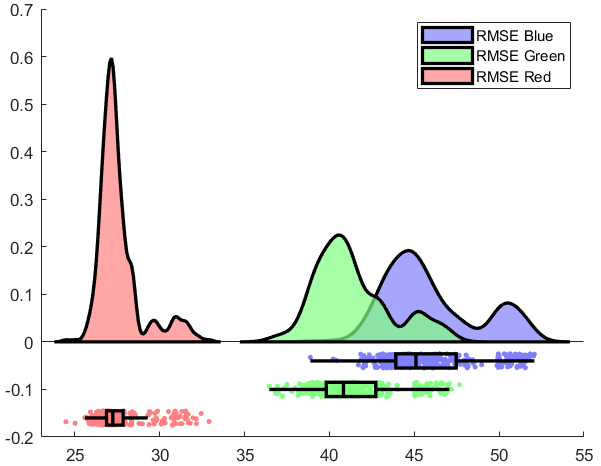}\\
		CP=17.73\%	&RMSE=33.95				& Avg. RMSE=38.33\\
	\end{tabular}
	\caption{The colorful board is scanned aided by dot patterns with different CPs. The temporal window size from top to bottom rows are respectively 2.5ms, 4.34ms, and 7.14ms.}
	\label{fig:ACSL_dots}
	\end{center}
\end{figure*}

Dot grid and circle patterns are one of the simplest patterns to detect the
local depth from SL~\cite{levy2019determining} or even calibrating the camera
when it is out of focus~\cite{wang2016accurate}. Changing the number of dots (feature
points) or their distance affects the depth resolution. However, more feature
points lead to additional processing time as well as generating more events,
which can lead to bus saturation in ECs. By changing the number of dots
dynamically based on the event rate, we are able to control the trade-off
between the speed of scanning and the amount of detail.
Fig.~\ref{fig:changing_SL}'s top two rows show the proposed ASL with dot-grid
patterns where $M$ and $N$ ($M<N$) are the number of dots on each grid.
Fig.~\ref{fig:ACSL_dots} shows three different dot patterns with different CPs.
The top row is generated with a temporal window size of $2.5ms$ (equivalent to
$400fps$). Similarly, the second row has a window of $4.34ms$ or $230fps$, and
the bottom row for $7.14ms$ or $140fps$. The leftmost column of
Fig.~\ref{fig:ACSL_dots} is a ground truth (GT) frame generated with a
one-second temporal window; the middle column is an example frame among the
$430$ frame samples. We compare each frame pixel by pixel with the GT frame to
compute the $RMSE$ for each channel, shown in the rightmost column.

\subsection{Multiple-lines pattern}
Since dot-grid patterns are leading to a sparse image, to generate a dense
image, line patterns are preferred in low-speed 3D scanning and multi-shot 3D measurement methods.
Sequential projection techniques mostly use strip lines~\cite{van2016real}. 
Since the DLP projector can quickly switch (4225 Hz) between
patterns, it is possible to generate a dense graph for some region of the object
by projecting lines and measuring the depth with triangulation. Although for the
spaces between lines we do not have measurements, increasing the number of
lines generates more features and it covers a larger area. Similarly to the
dot-grid pattern, increasing the number of lines or dots increases the scanning
processing time and event rate, and speed and detail must be traded off. The
third and fourth rows from top in Fig.~\ref{fig:changing_SL}, show the
proposed ASL with the line patterns where $M$ and $N$ ($M<N$) are the number of
lines in each pattern.
\begin{figure*}[htbp]
	\begin{center}
		\renewcommand{\tabcolsep}{1pt}
		\begin{tabular}{ccc}
			\includegraphics[width=0.32\textwidth]{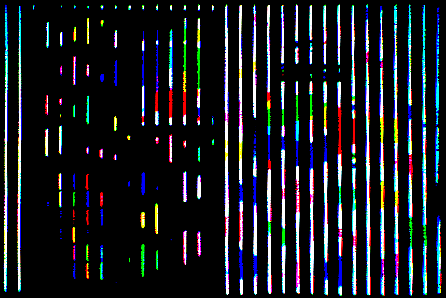}&
			\includegraphics[width=0.32\textwidth]{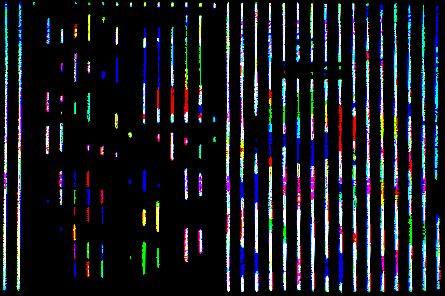}&
			\includegraphics[width=0.32\textwidth]{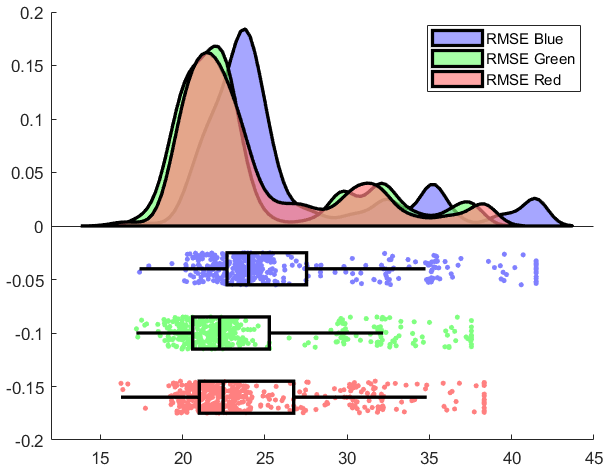}\\
			CP=7.02\%		&RMSE=17.54			& Avg. RMSE=25.00\\
			\hline
			\includegraphics[width=0.32\textwidth]{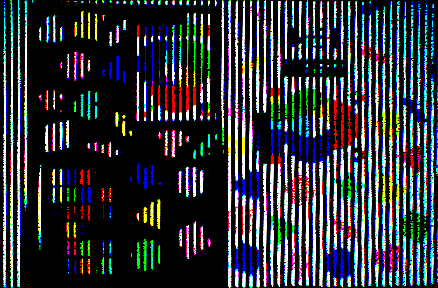}&
			\includegraphics[width=0.32\textwidth]{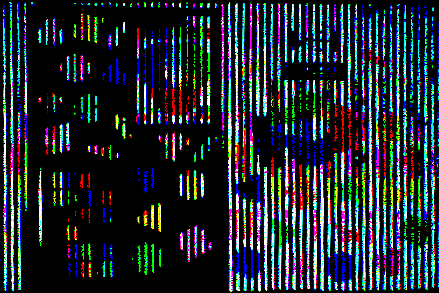}&
			\includegraphics[width=0.32\textwidth]{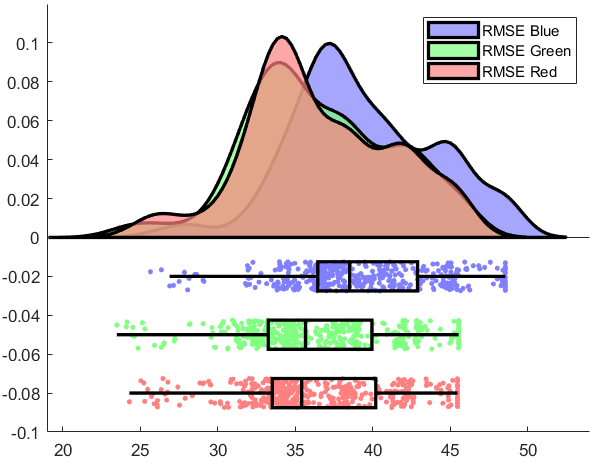}\\
			CP=14.04\%		&RMSE=25.37			& Avg. RMSE=37.48\\
			\hline
			\includegraphics[width=0.32\textwidth]{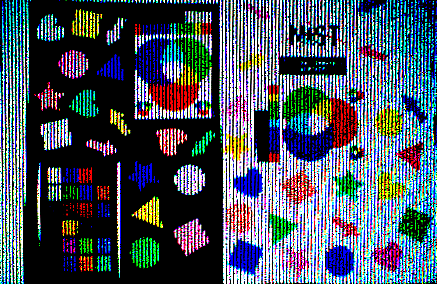}&
			\includegraphics[width=0.32\textwidth]{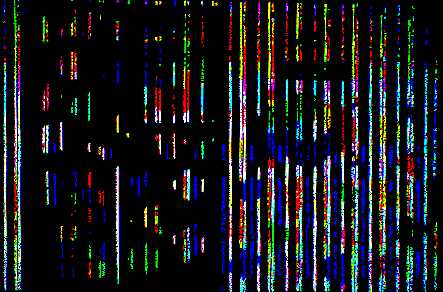}&
			\includegraphics[width=0.32\textwidth]{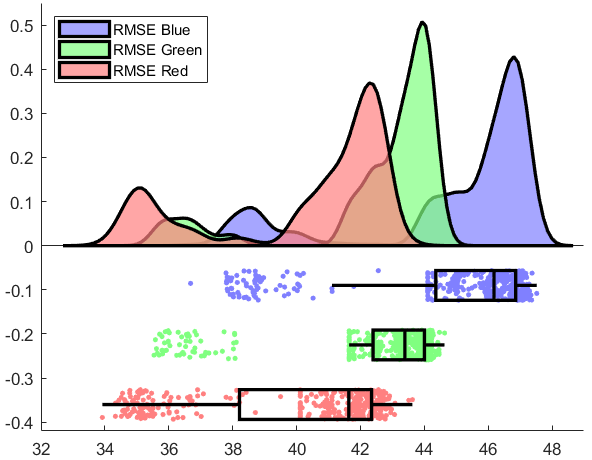}\\
			CP=28.07\%	&RMSE=36.69				& Avg. RMSE=42.55\\
		\end{tabular}
		\caption{The colorful board scanned by line patterns with different
      CPs. The temporal window size for the top row is 6.67ms and for the middle
      and the bottom rows is 7.14ms.}
		\label{fig:ACSL_lines}
	\end{center}
\end{figure*}

\subsection{Moving-line pattern}
\begin{figure}[htbp]
	\begin{center}
		\includegraphics[width=0.5\textwidth]{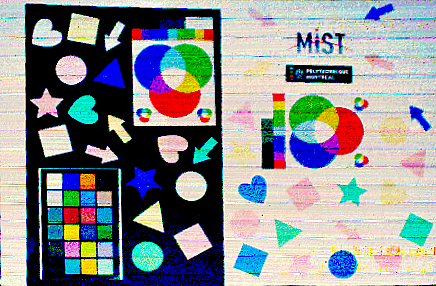}
		\caption{The colorful board scanned by moving-line pattern.}
		\label{fig:ACSL_moving-line}
	\end{center}
\end{figure}

To have a full dense scanning in 3D, a line pattern is very
common~\cite{brandli2014adaptive, matsuda2015mc3d, muglikar2021esl}. We propose
to use a moving line pattern (horizontal or vertical depending on the offset
between the camera and the projector), when the event rate is lower than the
bandwidth limits, providing dense scanning, as shown in the fifth row from top
in Fig.~\ref{fig:changing_SL}. Fig.~\ref{fig:ACSL_moving-line} shows the
colorful board scanned by a moving-line pattern.

\subsection{Solid pattern}
\begin{figure}[htbp]
	\begin{center}
		\includegraphics[width=1.1in]{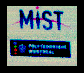}
		\caption{The colorful board scanned with a solid pattern for specific ROI. Note that the board was placed $160cm$ far from the monochrome EC that has a VGA resolution.}
		\label{fig:ACSL_solid-roi}
	\end{center}
\end{figure}

Whenever the 3D scanning is performed, or when we need the color information
only for a specific area (the region of interest), we can use the ROI mode. As
described in Section~\ref{acls}, Muglikar et al.~\cite{muglikar2021event} defined an
ROI dynamically based on the situation of the scene, then scanned that area with
more laser points. The bottom row of Fig.~\ref{fig:changing_SL}, indicates the
proposed ASL with the solid pattern for the ROI mode.
Fig.~\ref{fig:ACSL_solid-roi} shows the colorful board scanned by solid pattern
for a ROI that is the area of the MIST and Polytechnique Montreal logos.

\begin{figure*}[htbp]
	\begin{center}
		\includegraphics[width=0.49\textwidth]{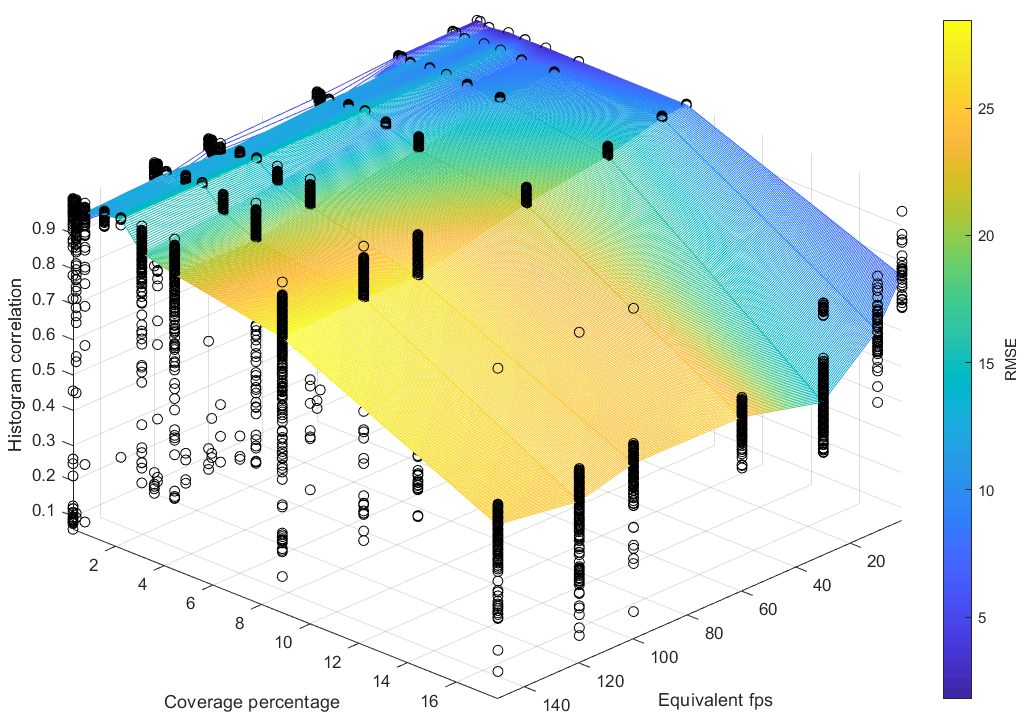}
		\includegraphics[width=0.49\textwidth]{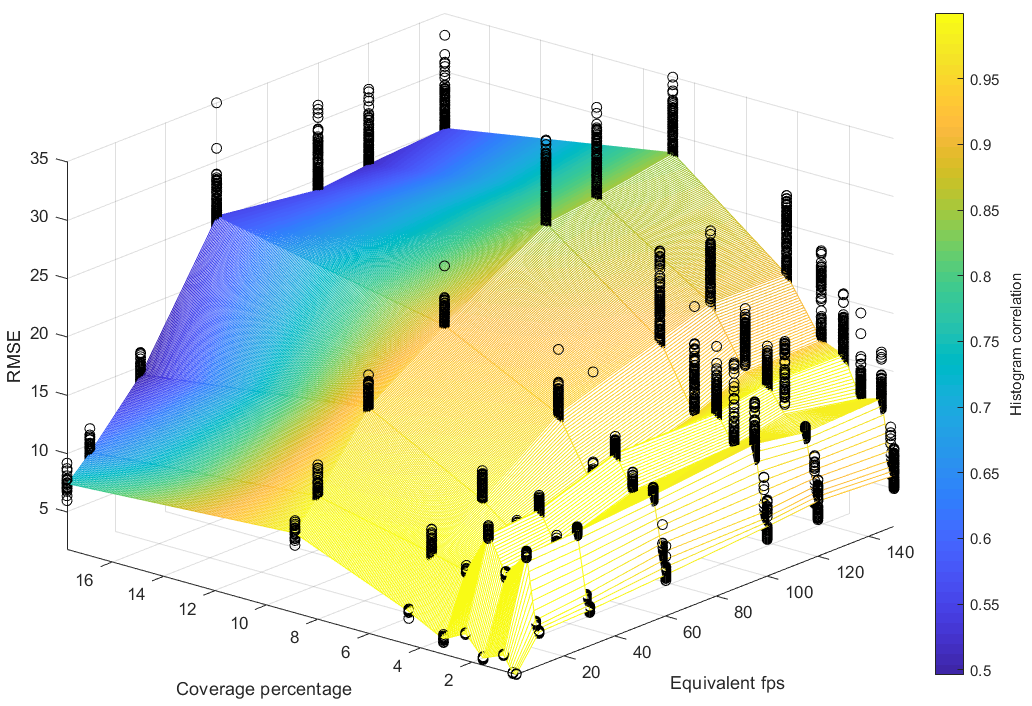}
		\caption{Comparing patterns with different CPs in speed and quality of the color detection.}
		\label{fig:ACSL_RMSE_HC}
	\end{center}
\end{figure*}

To compare different pattern and speed of scanning, we projected patterns with
various CPs onto the colorful board. Fig.~\ref{fig:ACSL_RMSE_HC} shows the
trade-off between details, speed, and the quality of the reconstructed colorful
image. It shows that to have a more detailed image, we need to spend more time
switching patterns to cover more area. Also, for high speed scanning, a sparse
pattern (lower CP with fewer details) is needed. It is worth noting that a
sparse pattern does not decrease the quality of the color detection even with
high speed sampling. Fig.~\ref{fig:ACSL_RMSE_HC}, has been generated by using
around 24000 frames.

\section{Conclusions}\label{conclusion}
We present a method to add color and depth to a monocular, monochrome
event-based camera while maintaining fast response time and resolution. Our
method reconstructs colorful events and frames using a monochrome EC aided by
adaptive structured light (ASL). By dynamically adjusting the projection, we have
color data when needed, managing the overall bandwidth of the system. We
achieved the color detection speed equivalent of 1400 fps with a Texas
Instrument's DLP LightCrafter 4500 projector. Our method could be used in
event-based depth measurement and perception projects. Advantages of ECs, makes
the colorful depth detection much faster than RGBD cameras.

Although the color detection is related to the lighting conditions and material
properties at the intersection point (object surface), the scope of this work
was the color detection on common materials that are generally matte and not too
shiny (with high reflection) or fluorescence. Some materials can interact with
light: they can be absorbing, scattering or emitting
light~\cite{guilbault2020practical}. In this work we focused on visual light
wavelength (emitted by the LED projector) and materials that are not in the
category of fluorescence and they do not change the wavelength of the light.
However, the use of the event-based camera with a different type of light source
and materials could be investigated in future works. Also, without considering
color detection, static reflective materials can be scanned more effectively
with ECs when compared to the other depth measurement devices~\cite{matsuda2015mc3d}.
To detect the color of these kind of materials, a Blinn-Phong shading
model~\cite{blinn1977models} could be considered in future works.

\bibliographystyle{IEEEtran}

\begin{thebibliography}{10}
\providecommand{\url}[1]{#1}
\csname url@samestyle\endcsname
\providecommand{\newblock}{\relax}
\providecommand{\bibinfo}[2]{#2}
\providecommand{\BIBentrySTDinterwordspacing}{\spaceskip=0pt\relax}
\providecommand{\BIBentryALTinterwordstretchfactor}{4}
\providecommand{\BIBentryALTinterwordspacing}{\spaceskip=\fontdimen2\font plus
\BIBentryALTinterwordstretchfactor\fontdimen3\font minus
  \fontdimen4\font\relax}
\providecommand{\BIBforeignlanguage}[2]{{%
\expandafter\ifx\csname l@#1\endcsname\relax
\typeout{** WARNING: IEEEtran.bst: No hyphenation pattern has been}%
\typeout{** loaded for the language `#1'. Using the pattern for}%
\typeout{** the default language instead.}%
\else
\language=\csname l@#1\endcsname
\fi
#2}}
\providecommand{\BIBdecl}{\relax}
\BIBdecl

\bibitem{gallego2020event}
\BIBentryALTinterwordspacing
G.~Gallego, T.~Delbruck, G.~Orchard, C.~Bartolozzi, B.~Taba, A.~Censi,
  S.~Leutenegger, A.~Davison, J.~Conradt, K.~Daniilidis \emph{et~al.},
  ``Event-based vision: A survey,'' \emph{IEEE Transactions on Pattern Analysis
  and Machine Intelligence}, 2020. [Online]. Available:
  \url{https://ieeexplore.ieee.org/document/9138762}
\BIBentrySTDinterwordspacing

\bibitem{barrios2018movement}
\BIBentryALTinterwordspacing
J.~Barrios-Avil{\'e}s, T.~Iakymchuk, J.~Samaniego, L.~D. Medus, and
  A.~Rosado-Mu{\~n}oz, ``Movement detection with event-based cameras:
  Comparison with frame-based cameras in robot object tracking using powerlink
  communication,'' \emph{Electronics}, vol.~7, no.~11, p. 304, 2018. [Online].
  Available: \url{https://www.mdpi.com/2079-9292/7/11/304}
\BIBentrySTDinterwordspacing

\bibitem{mitrokhin2018event}
\BIBentryALTinterwordspacing
A.~Mitrokhin, C.~Ferm{\"u}ller, C.~Parameshwara, and Y.~Aloimonos,
  ``Event-based moving object detection and tracking,'' in \emph{2018 IEEE/RSJ
  International Conference on Intelligent Robots and Systems (IROS)}.\hskip 1em
  plus 0.5em minus 0.4em\relax IEEE, 2018, pp. 1--9. [Online]. Available:
  \url{https://ieeexplore.ieee.org/document/8593805}
\BIBentrySTDinterwordspacing

\bibitem{alzugaray2018asynchronous}
\BIBentryALTinterwordspacing
I.~Alzugaray and M.~Chli, ``Asynchronous corner detection and tracking for
  event cameras in real time,'' \emph{IEEE Robotics and Automation Letters},
  vol.~3, no.~4, pp. 3177--3184, 2018. [Online]. Available:
  \url{https://ieeexplore.ieee.org/document/8392795}
\BIBentrySTDinterwordspacing

\bibitem{zihao2017event}
\BIBentryALTinterwordspacing
A.~Zihao~Zhu, N.~Atanasov, and K.~Daniilidis, ``Event-based visual inertial
  odometry,'' in \emph{Proceedings of the IEEE Conference on Computer Vision
  and Pattern Recognition}, 2017, pp. 5391--5399. [Online]. Available:
  \url{https://openaccess.thecvf.com/content_cvpr_2017/html/Zhu_Event-Based_Visual_Inertial_CVPR_2017_paper.html}
\BIBentrySTDinterwordspacing

\bibitem{mueggler2018continuous}
\BIBentryALTinterwordspacing
E.~Mueggler, G.~Gallego, H.~Rebecq, and D.~Scaramuzza, ``Continuous-time
  visual-inertial odometry for event cameras,'' \emph{IEEE Transactions on
  Robotics}, vol.~34, no.~6, pp. 1425--1440, 2018. [Online]. Available:
  \url{https://ieeexplore.ieee.org/document/8432102}
\BIBentrySTDinterwordspacing

\bibitem{rebecq2016evo}
\BIBentryALTinterwordspacing
H.~Rebecq, T.~Horstsch{\"a}fer, G.~Gallego, and D.~Scaramuzza, ``{EVO}: A
  geometric approach to event-based 6-dof parallel tracking and mapping in real
  time,'' \emph{IEEE Robotics and Automation Letters}, vol.~2, no.~2, pp.
  593--600, 2017. [Online]. Available:
  \url{https://ieeexplore.ieee.org/document/7797445}
\BIBentrySTDinterwordspacing

\bibitem{reinbacher2017real}
\BIBentryALTinterwordspacing
C.~Reinbacher, G.~Munda, and T.~Pock, ``Real-time panoramic tracking for event
  cameras,'' in \emph{2017 IEEE International Conference on Computational
  Photography (ICCP)}.\hskip 1em plus 0.5em minus 0.4em\relax IEEE, 2017, pp.
  1--9. [Online]. Available: \url{https://ieeexplore.ieee.org/document/7951488}
\BIBentrySTDinterwordspacing

\bibitem{sironi2018hats}
\BIBentryALTinterwordspacing
A.~Sironi, M.~Brambilla, N.~Bourdis, X.~Lagorce, and R.~Benosman, ``Hats:
  Histograms of averaged time surfaces for robust event-based object
  classification,'' in \emph{Proceedings of the IEEE Conference on Computer
  Vision and Pattern Recognition}, 2018, pp. 1731--1740. [Online]. Available:
  \url{https://openaccess.thecvf.com/content_cvpr_2018/html/Sironi_HATS_Histograms_of_CVPR_2018_paper.html}
\BIBentrySTDinterwordspacing

\bibitem{zhou2018semi}
\BIBentryALTinterwordspacing
Y.~Zhou, G.~Gallego, H.~Rebecq, L.~Kneip, H.~Li, and D.~Scaramuzza,
  ``Semi-dense 3d reconstruction with a stereo event camera,'' in
  \emph{Proceedings of the European Conference on Computer Vision
  (ECCV)}.\hskip 1em plus 0.5em minus 0.4em\relax Springer International
  Publishing, 2018, pp. 235--251. [Online]. Available:
  \url{https://link.springer.com/chapter/10.1007}
\BIBentrySTDinterwordspacing

\bibitem{rebecq2018emvs}
\BIBentryALTinterwordspacing
H.~Rebecq, G.~Gallego, E.~Mueggler, and D.~Scaramuzza, ``{EMVS: Event-based
  multi-view stereo—3D reconstruction with an event camera in real-time},''
  \emph{International Journal of Computer Vision}, vol. 126, no.~12, pp.
  1394--1414, 2018. [Online]. Available:
  \url{https://link.springer.com/article/10.1007}
\BIBentrySTDinterwordspacing

\bibitem{steffen2019neuromorphic}
\BIBentryALTinterwordspacing
L.~Steffen, D.~Reichard, J.~Weinland, J.~Kaiser, A.~Roennau, and R.~Dillmann,
  ``Neuromorphic stereo vision: A survey of bio-inspired sensors and
  algorithms,'' \emph{Frontiers in Neurorobotics}, vol.~13, p.~28, 2019.
  [Online]. Available:
  \url{https://www.frontiersin.org/articles/10.3389/fnbot.2019.00028/full}
\BIBentrySTDinterwordspacing

\bibitem{tremeau2008color}
\BIBentryALTinterwordspacing
A.~Tremeau, S.~Tominaga, and K.~Plataniotis, ``Color in image and video
  processing: most recent trends and future research directions,''
  \emph{EURASIP Journal on Image and Video Processing}, vol. 2008, pp. 1--26,
  2008. [Online]. Available:
  \url{https://link.springer.com/content/pdf/10.1155/2008/581371.pdf}
\BIBentrySTDinterwordspacing

\bibitem{marcireau2018event}
\BIBentryALTinterwordspacing
A.~Marcireau, S.-H. Ieng, C.~Simon-Chane, and R.~B. Benosman, ``Event-based
  color segmentation with a high dynamic range sensor,'' \emph{Frontiers in
  neuroscience}, vol.~12, p. 135, 2018. [Online]. Available:
  \url{https://www.frontiersin.org/articles/10.3389/fnins.2018.00135/pdf}
\BIBentrySTDinterwordspacing

\bibitem{li2015design}
\BIBentryALTinterwordspacing
C.~Li, C.~Brandli, R.~Berner, H.~Liu, M.~Yang, S.-C. Liu, and T.~Delbruck,
  ``Design of an rgbw color vga rolling and global shutter dynamic and
  active-pixel vision sensor,'' in \emph{2015 IEEE International Symposium on
  Circuits and Systems (ISCAS)}.\hskip 1em plus 0.5em minus 0.4em\relax IEEE,
  2015, pp. 718--721. [Online]. Available:
  \url{https://ieeexplore.ieee.org/document/7168734}
\BIBentrySTDinterwordspacing

\bibitem{taverni2018front}
\BIBentryALTinterwordspacing
G.~Taverni, D.~P. Moeys, C.~Li, C.~Cavaco, V.~Motsnyi, D.~S.~S. Bello, and
  T.~Delbruck, ``Front and back illuminated dynamic and active pixel vision
  sensors comparison,'' \emph{IEEE Transactions on Circuits and Systems II:
  Express Briefs}, vol.~65, no.~5, pp. 677--681, 2018. [Online]. Available:
  \url{https://ieeexplore.ieee.org/document/8334288}
\BIBentrySTDinterwordspacing

\bibitem{moeys2017color}
\BIBentryALTinterwordspacing
D.~P. Moeys, C.~Li, J.~N. Martel, S.~Bamford, L.~Longinotti, V.~Motsnyi,
  D.~S.~S. Bello, and T.~Delbruck, ``Color temporal contrast sensitivity in
  dynamic vision sensors,'' in \emph{2017 IEEE International Symposium on
  Circuits and Systems (ISCAS)}.\hskip 1em plus 0.5em minus 0.4em\relax IEEE,
  2017, pp. 1--4. [Online]. Available:
  \url{https://ieeexplore.ieee.org/document/8050412}
\BIBentrySTDinterwordspacing

\bibitem{moeys2017sensitive}
\BIBentryALTinterwordspacing
D.~P. Moeys, F.~Corradi, C.~Li, S.~A. Bamford, L.~Longinotti, F.~F. Voigt,
  S.~Berry, G.~Taverni, F.~Helmchen, and T.~Delbruck, ``A sensitive dynamic and
  active pixel vision sensor for color or neural imaging applications,''
  \emph{IEEE transactions on biomedical circuits and systems}, vol.~12, no.~1,
  pp. 123--136, 2017. [Online]. Available:
  \url{https://ieeexplore.ieee.org/document/8094907}
\BIBentrySTDinterwordspacing

\bibitem{brandli2014adaptive}
\BIBentryALTinterwordspacing
C.~Brandli, T.~Mantel, M.~Hutter, M.~H{\"o}pflinger, R.~Berner, R.~Siegwart,
  and T.~Delbruck, ``Adaptive pulsed laser line extraction for terrain
  reconstruction using a dynamic vision sensor,'' \emph{Frontiers in
  neuroscience}, vol.~7, p. 275, 2014. [Online]. Available:
  \url{https://www.frontiersin.org/articles/10.3389/fnins.2013.00275/full}
\BIBentrySTDinterwordspacing

\bibitem{matsuda2015mc3d}
\BIBentryALTinterwordspacing
N.~Matsuda, O.~Cossairt, and M.~Gupta, ``{MC3D: Motion Contrast 3D Scanning},''
  in \emph{2015 IEEE International Conference on Computational Photography
  (ICCP)}.\hskip 1em plus 0.5em minus 0.4em\relax IEEE, 2015, pp. 1--10.
  [Online]. Available: \url{https://ieeexplore.ieee.org/document/7168370}
\BIBentrySTDinterwordspacing

\bibitem{muglikar2021event}
\BIBentryALTinterwordspacing
M.~Muglikar, D.~P. Moeys, and D.~Scaramuzza, ``Event guided depth sensing,'' in
  \emph{2021 International Conference on 3D Vision (3DV)}.\hskip 1em plus 0.5em
  minus 0.4em\relax IEEE, 2021, pp. 385--393. [Online]. Available:
  \url{https://ieeexplore.ieee.org/document/9665844}
\BIBentrySTDinterwordspacing

\bibitem{muglikar2021esl}
\BIBentryALTinterwordspacing
M.~Muglikar, G.~Gallego, and D.~Scaramuzza, ``Esl: Event-based structured
  light,'' in \emph{2021 International Conference on 3D Vision (3DV)}.\hskip
  1em plus 0.5em minus 0.4em\relax IEEE, 2021, pp. 1165--1174. [Online].
  Available: \url{https://ieeexplore.ieee.org/document/9665929}
\BIBentrySTDinterwordspacing

\bibitem{bayer1976color}
B.~E. Bayer, ``Color imaging array,'' \emph{United States Patent 3,971,065},
  1976.

\bibitem{ramanath2005color}
\BIBentryALTinterwordspacing
R.~Ramanath, W.~E. Snyder, Y.~Yoo, and M.~S. Drew, ``Color image processing
  pipeline,'' \emph{IEEE Signal Processing Magazine}, vol.~22, no.~1, pp.
  34--43, 2005. [Online]. Available:
  \url{https://ieeexplore.ieee.org/document/1407713}
\BIBentrySTDinterwordspacing

\bibitem{khashabi2014joint}
\BIBentryALTinterwordspacing
D.~Khashabi, S.~Nowozin, J.~Jancsary, and A.~W. Fitzgibbon, ``Joint demosaicing
  and denoising via learned nonparametric random fields,'' \emph{IEEE
  Transactions on Image Processing}, vol.~23, no.~12, pp. 4968--4981, 2014.
  [Online]. Available: \url{https://ieeexplore.ieee.org/document/6906294}
\BIBentrySTDinterwordspacing

\bibitem{levin2004colorization}
\BIBentryALTinterwordspacing
A.~Levin, D.~Lischinski, and Y.~Weiss, ``Colorization using optimization,'' in
  \emph{ACM SIGGRAPH 2004 Papers}, 2004, pp. 689--694. [Online]. Available:
  \url{https://dl.acm.org/doi/10.1145/1186562.1015780}
\BIBentrySTDinterwordspacing

\bibitem{zhang2016colorful}
\BIBentryALTinterwordspacing
R.~Zhang, P.~Isola, and A.~A. Efros, ``Colorful image colorization,'' in
  \emph{European conference on computer vision}.\hskip 1em plus 0.5em minus
  0.4em\relax Springer, 2016, pp. 649--666. [Online]. Available:
  \url{https://link.springer.com/chapter/10.1007/978-3-319-46487-9_40}
\BIBentrySTDinterwordspacing

\bibitem{jang2020deep}
\BIBentryALTinterwordspacing
H.~W. Jang and Y.~J. Jung, ``Deep color transfer for color-plus-mono dual
  cameras,'' \emph{Sensors}, vol.~20, no.~9, p. 2743, 2020. [Online].
  Available: \url{https://www.mdpi.com/1424-8220/20/9/2743}
\BIBentrySTDinterwordspacing

\bibitem{munda2018real}
\BIBentryALTinterwordspacing
G.~Munda, C.~Reinbacher, and T.~Pock, ``Real-time intensity-image
  reconstruction for event cameras using manifold regularisation,''
  \emph{International Journal of Computer Vision}, vol. 126, no.~12, pp.
  1381--1393, 2018. [Online]. Available:
  \url{https://link.springer.com/article/10.1007/s11263-018-1106-2}
\BIBentrySTDinterwordspacing

\bibitem{scheerlinck2018continuous}
\BIBentryALTinterwordspacing
C.~Scheerlinck, N.~Barnes, and R.~Mahony, ``Continuous-time intensity
  estimation using event cameras,'' in \emph{Asian Conference on Computer
  Vision}.\hskip 1em plus 0.5em minus 0.4em\relax Springer, 2018, pp. 308--324.
  [Online]. Available:
  \url{https://link.springer.com/chapter/10.1007/978-3-030-20873-8_20}
\BIBentrySTDinterwordspacing

\bibitem{rebecq2019events}
\BIBentryALTinterwordspacing
H.~Rebecq, R.~Ranftl, V.~Koltun, and D.~Scaramuzza, ``Events-to-video: Bringing
  modern computer vision to event cameras,'' in \emph{Proceedings of the
  IEEE/CVF Conference on Computer Vision and Pattern Recognition}, 2019, pp.
  3857--3866. [Online]. Available:
  \url{https://openaccess.thecvf.com/content_CVPR_2019/html/Rebecq_Events-To-Video_Bringing_Modern_Computer_Vision_to_Event_Cameras_CVPR_2019_paper.html}
\BIBentrySTDinterwordspacing

\bibitem{scheerlinck2019asynchronous}
\BIBentryALTinterwordspacing
C.~Scheerlinck, N.~Barnes, and R.~Mahony, ``Asynchronous spatial image
  convolutions for event cameras,'' \emph{IEEE Robotics and Automation
  Letters}, vol.~4, no.~2, pp. 816--822, 2019. [Online]. Available:
  \url{https://ieeexplore.ieee.org/document/8613800}
\BIBentrySTDinterwordspacing

\bibitem{pan2019bringing}
\BIBentryALTinterwordspacing
L.~Pan, C.~Scheerlinck, X.~Yu, R.~Hartley, M.~Liu, and Y.~Dai, ``Bringing a
  blurry frame alive at high frame-rate with an event camera,'' in
  \emph{Proceedings of the IEEE/CVF Conference on Computer Vision and Pattern
  Recognition}, 2019, pp. 6820--6829. [Online]. Available:
  \url{https://openaccess.thecvf.com/content_CVPR_2019/html/Pan_Bringing_a_Blurry_Frame_Alive_at_High_Frame-Rate_With_an_CVPR_2019_paper.html}
\BIBentrySTDinterwordspacing

\bibitem{haoyu2020learning}
\BIBentryALTinterwordspacing
C.~Haoyu, T.~Minggui, S.~Boxin, W.~YIzhou, and H.~Tiejun, ``Learning to deblur
  and generate high frame rate video with an event camera,'' \emph{arXiv
  preprint arXiv:2003.00847}, 2020. [Online]. Available:
  \url{https://arxiv.org/abs/2003.00847}
\BIBentrySTDinterwordspacing

\bibitem{messikommer2022multi}
\BIBentryALTinterwordspacing
N.~Messikommer, S.~Georgoulis, D.~Gehrig, S.~Tulyakov, J.~Erbach,
  A.~Bochicchio, Y.~Li, and D.~Scaramuzza, ``Multi-bracket high dynamic range
  imaging with event cameras,'' \emph{arXiv preprint arXiv:2203.06622}, 2022.
  [Online]. Available: \url{https://arxiv.org/abs/2203.06622}
\BIBentrySTDinterwordspacing

\bibitem{rebecq2019high}
\BIBentryALTinterwordspacing
H.~Rebecq, R.~Ranftl, V.~Koltun, and D.~Scaramuzza, ``High speed and high
  dynamic range video with an event camera,'' \emph{IEEE transactions on
  pattern analysis and machine intelligence}, vol.~43, no.~6, pp. 1964--1980,
  2019. [Online]. Available: \url{https://ieeexplore.ieee.org/document/8946715}
\BIBentrySTDinterwordspacing

\bibitem{pan2020high}
\BIBentryALTinterwordspacing
L.~Pan, R.~Hartley, C.~Scheerlinck, M.~Liu, X.~Yu, and Y.~Dai, ``High frame
  rate video reconstruction based on an event camera,'' \emph{IEEE Transactions
  on Pattern Analysis and Machine Intelligence}, 2020. [Online]. Available:
  \url{https://ieeexplore.ieee.org/document/9252186}
\BIBentrySTDinterwordspacing

\bibitem{mostafavi2021learning}
\BIBentryALTinterwordspacing
M.~Mostafavi, L.~Wang, and K.-J. Yoon, ``Learning to reconstruct hdr images
  from events, with applications to depth and flow prediction,''
  \emph{International Journal of Computer Vision}, vol. 129, no.~4, pp.
  900--920, 2021. [Online]. Available:
  \url{https://link.springer.com/article/10.1007/s11263-020-01410-2}
\BIBentrySTDinterwordspacing

\bibitem{scheerlinck2019ced}
\BIBentryALTinterwordspacing
C.~Scheerlinck, H.~Rebecq, T.~Stoffregen, N.~Barnes, R.~Mahony, and
  D.~Scaramuzza, ``Ced: Color event camera dataset,'' in \emph{Proceedings of
  the IEEE/CVF Conference on Computer Vision and Pattern Recognition
  Workshops}, 2019, pp. 0--0. [Online]. Available:
  \url{https://openaccess.thecvf.com/content_CVPRW_2019/html/EventVision/Scheerlinck_CED_Color_Event_Camera_Dataset_CVPRW_2019_paper.html}
\BIBentrySTDinterwordspacing

\bibitem{gehrig2022high}
\BIBentryALTinterwordspacing
D.~Gehrig and D.~Scaramuzza, ``Are high-resolution event cameras really
  needed?'' \emph{arXiv preprint arXiv:2203.14672}, 2022. [Online]. Available:
  \url{https://arxiv.org/abs/2203.14672}
\BIBentrySTDinterwordspacing

\bibitem{muthu2002red}
\BIBentryALTinterwordspacing
S.~Muthu, F.~J. Schuurmans, and M.~D. Pashley, ``Red, green, and blue led based
  white light generation: issues and control,'' in \emph{Conference Record of
  the 2002 IEEE Industry Applications Conference. 37th IAS Annual Meeting (Cat.
  No. 02CH37344)}, vol.~1.\hskip 1em plus 0.5em minus 0.4em\relax IEEE, 2002,
  pp. 327--333. [Online]. Available:
  \url{https://ieeexplore.ieee.org/document/1044108}
\BIBentrySTDinterwordspacing

\bibitem{muthu2003red}
\BIBentryALTinterwordspacing
S.~Muthu and J.~Gaines, ``Red, green and blue led-based white light source:
  implementation challenges and control design,'' in \emph{38th IAS Annual
  Meeting on Conference Record of the Industry Applications Conference, 2003.},
  vol.~1.\hskip 1em plus 0.5em minus 0.4em\relax IEEE, 2003, pp. 515--522.
  [Online]. Available: \url{https://ieeexplore.ieee.org/document/1257549}
\BIBentrySTDinterwordspacing

\bibitem{david2018led}
\BIBentryALTinterwordspacing
A.~David and L.~A. Whitehead, ``Led-based white light,'' \emph{Comptes Rendus
  Physique}, vol.~19, no.~3, pp. 169--181, 2018. [Online]. Available:
  \url{https://www.sciencedirect.com/science/article/pii/S163107051830029X}
\BIBentrySTDinterwordspacing

\bibitem{pharr2016physically}
\BIBentryALTinterwordspacing
M.~Pharr, W.~Jakob, and G.~Humphreys, \emph{Physically based rendering: From
  theory to implementation}.\hskip 1em plus 0.5em minus 0.4em\relax Morgan
  Kaufmann, 2016. [Online]. Available:
  \url{https://www.sciencedirect.com/book/9780128006450/physically-based-rendering}
\BIBentrySTDinterwordspacing

\bibitem{choi2020color}
\BIBentryALTinterwordspacing
W.~Choi, H.~S. Park, and C.-M. Kyung, ``Color reproduction pipeline for an rgbw
  color filter array sensor,'' \emph{Optics Express}, vol.~28, no.~10, pp.
  15\,678--15\,690, 2020. [Online]. Available:
  \url{https://opg.optica.org/oe/fulltext.cfm?uri=oe-28-10-15678&id=431622}
\BIBentrySTDinterwordspacing

\bibitem{finateu20205}
\BIBentryALTinterwordspacing
T.~Finateu, A.~Niwa, D.~Matolin, K.~Tsuchimoto, A.~Mascheroni, E.~Reynaud,
  P.~Mostafalu, F.~Brady, L.~Chotard, F.~LeGoff \emph{et~al.}, ``5.10 a
  1280$\times$ 720 back-illuminated stacked temporal contrast event-based
  vision sensor with 4.86 $\mu$m pixels, 1.066 geps readout, programmable
  event-rate controller and compressive data-formatting pipeline,'' in
  \emph{2020 IEEE International Solid-State Circuits Conference-(ISSCC)}.\hskip
  1em plus 0.5em minus 0.4em\relax IEEE, 2020, pp. 112--114. [Online].
  Available: \url{https://ieeexplore.ieee.org/document/9063149}
\BIBentrySTDinterwordspacing

\bibitem{delbruck2021feedback}
\BIBentryALTinterwordspacing
T.~Delbruck, R.~Graca, and M.~Paluch, ``Feedback control of event cameras,'' in
  \emph{Proceedings of the IEEE/CVF Conference on Computer Vision and Pattern
  Recognition}, 2021, pp. 1324--1332. [Online]. Available:
  \url{https://openaccess.thecvf.com/content/CVPR2021W/EventVision/html/Delbruck_Feedback_Control_of_Event_Cameras_CVPRW_2021_paper.html}
\BIBentrySTDinterwordspacing

\bibitem{levy2019determining}
\BIBentryALTinterwordspacing
H.~Levy, ``Determining local depth from structured light using a regular dot
  grid,'' \emph{Technical Disclosure Commons}, 2019. [Online]. Available:
  \url{https://www.tdcommons.org/dpubs_series/2536/}
\BIBentrySTDinterwordspacing

\bibitem{wang2016accurate}
\BIBentryALTinterwordspacing
Y.~Wang, X.~Chen, J.~Tao, K.~Wang, and M.~Ma, ``Accurate feature detection for
  out-of-focus camera calibration,'' \emph{Applied optics}, vol.~55, no.~28,
  pp. 7964--7971, 2016. [Online]. Available:
  \url{https://opg.optica.org/ao/fulltext.cfm?uri=ao-55-28-7964&id=350405}
\BIBentrySTDinterwordspacing

\bibitem{van2016real}
\BIBentryALTinterwordspacing
S.~Van~der Jeught and J.~J. Dirckx, ``Real-time structured light profilometry:
  a review,'' \emph{Optics and Lasers in Engineering}, vol.~87, pp. 18--31,
  2016. [Online]. Available:
  \url{https://www.sciencedirect.com/science/article/pii/S0143816616000166}
\BIBentrySTDinterwordspacing

\bibitem{guilbault2020practical}
G.~G. Guilbault, \emph{Practical fluorescence}.\hskip 1em plus 0.5em minus
  0.4em\relax CRC Press, 2020.

\bibitem{blinn1977models}
\BIBentryALTinterwordspacing
J.~F. Blinn, ``Models of light reflection for computer synthesized pictures,''
  in \emph{Proceedings of the 4th annual conference on Computer graphics and
  interactive techniques}, 1977, pp. 192--198. [Online]. Available:
  \url{https://dl.acm.org/doi/abs/10.1145/563858.563893}
\BIBentrySTDinterwordspacing

\end{thebibliography}

\end{document}